\documentclass[letterpaper, 10 pt, conference]{ieeeconf}  % Comment this line out if you need a4paper

\IEEEoverridecommandlockouts                              % This command is only needed if 
\usepackage{times}
\usepackage{epsfig}
\usepackage{graphicx}
\usepackage{amsmath}
\usepackage{amssymb}
\usepackage{soul}
\usepackage[utf8]{inputenc} %para tildes 
\usepackage[disable]{todonotes}	%[disable] para desactivar
\usepackage{color}	%para usar azul para comentarios
\usepackage{adjustbox}
\usepackage{multirow}
\usepackage{rotating}

\usepackage[caption=false,font=footnotesize]{subfig}
\newcommand{\citep}{\cite}

\newcommand\norm[1]{\left\lVert#1\right\rVert}

\newcommand{\etal}[1]{\textit{et al. }}
\newcommand{\fig}[1]{Fig.~\ref{#1}}
\newcommand{\tab}[1]{Table~\ref{#1}}

\title{\LARGE \bf
Scalable Active Learning for Object Detection
}

\author{Elmar Haussmann, Michele Fenzi, Kashyap Chitta, Jan Ivanecky, Hanson Xu, Donna Roy\\ Akshita Mittel, Nicolas Koumchatzky, Clement Farabet, Jose M. Alvarez% <-this % stops a space
\thanks{Elmar Haussmann and Michele Fenzi are with NVIDIA, Germany. Hanson Xu is with NVIDIA, China. Jan Ivanecky, Donna Roy, Akshita Mittel, Nicolas Koumchatzky, Clement Farabet and Jose M. Alvarez are with NVIDIA, USA.   Contact: {\tt\{ehaussmann,mfenzi,josea,cfarabet\}@nvidia.com}}%
\thanks{Kashyap Chitta is with the Max Planck Institute for Intelligent Systems, Tübingen (Germany); Work was done while he was an intern at NVIDIA. %{\tt\small josea@nvidia.com}
}
\thanks{Special thanks to the NVIDIA MagLev Infrastructure team for their support and invaluable help.}
 }
\begin{document}

\maketitle
\thispagestyle{empty}
\pagestyle{empty}

%%%%%%%%%%%%%%%%%%%%%%%%%%%%%%%%%%%%%%%%%%%%%%%%%%%%%%%%%%%%%%%%%%%%%%%%%%%%%%%%
\begin{abstract}
%Active Learning is a powerful technique to improve data efficiency for supervised learning methods. In short, AL aims to select, from a large unlabeled dataset, the smallest possible training set to solve a specific task. We have built a scalable production system for Active Learning in the domain of Autonomous driving. In this paper, we describe the resulting high-level design, sketch some of the challenges and their solutions, our current results at scale and briefly describe the open problems and future directions.
Deep Neural Networks trained in a fully supervised fashion are the dominant technology in perception-based autonomous driving systems. While collecting large amounts of unlabeled data is already a major undertaking, only a subset of it can be labeled by humans due to the effort needed for high-quality annotation. Therefore, finding the right data to label has become a key challenge. Active learning is a powerful technique to improve data efficiency for supervised learning methods, as it aims at selecting the smallest possible training set to reach a required performance. We have built a scalable production system for active learning in the domain of autonomous driving. In this paper, we describe the resulting high-level design, sketch some of the challenges and their solutions, present our current results at scale, and briefly describe the open problems and future directions.
\end{abstract}
%\vspace{5pt}

\section{Introduction}
Deep Neural Networks (DNNs) trained in a fully supervised fashion are the dominant technology in perception-based autonomous driving systems. The performance of these DNNs hinges on the amount and quality of data used to train them. Therefore, having a large and diverse training dataset that covers all relevant scenarios is key to achieve the accuracy implied by safety-critical operation.

\begin{figure}[!t]
\centering
\includegraphics[width=0.99\linewidth] {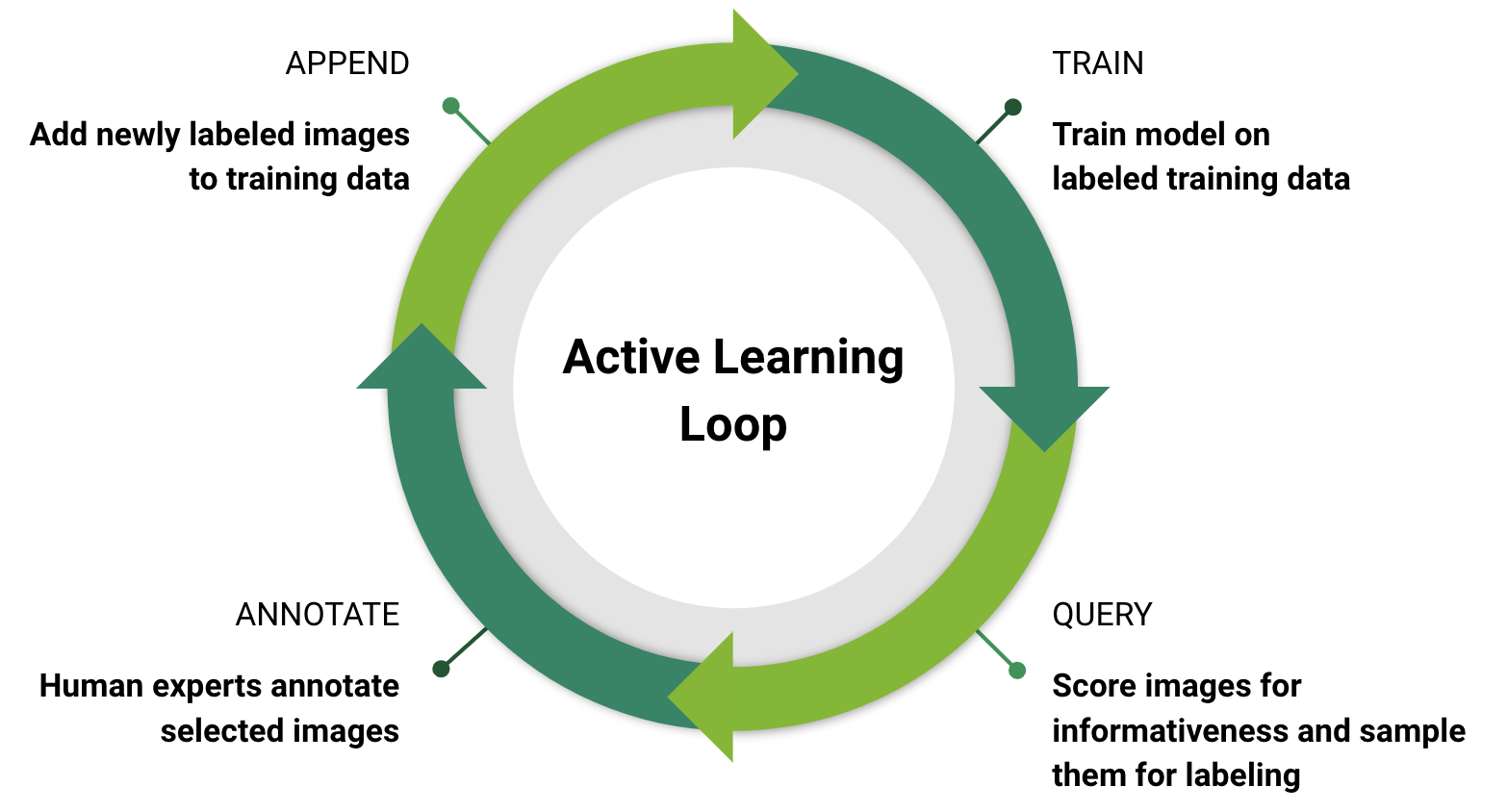}
\vspace{-0.3cm}
\caption{Active learning loop diagram. At each iteration, the scoring function and sampling strategy in the \textit{query} step decide which images should be sent to labeling and added to the training dataset for further training.}
\label{fig:Active_Learning}
\vspace{-0.5cm}
\end{figure}

Active learning is a powerful technique to improve data efficiency for supervised learning methods. The key idea behind active learning is that a machine learning algorithm can achieve greater accuracy with fewer training labels if it is allowed to choose the data from which it learns \cite{Settles2010Active}.
Figure~\ref{fig:Active_Learning} shows an outline of an active learning loop. In each iteration, the size of the dataset used to train a machine learning model is increased by labeling new data selected from a pool of unlabeled data. Crucially, in the \textit{query phase}, the model is ``actively" involved in choosing which images to label, e.g., by selecting images that cause the highest uncertainty. This has two potential advantages. First, automation: the decision about which images to label would otherwise need to be (largely) done manually, which can be labor intensive. Second, performance: by involving the model in building the training dataset, it may be possible to achieve higher performance with fewer data samples, e.g., the model may choose images it learns most from-- something often difficult for a human.

While there have been emerging efforts to improve deep active learning for classification~\cite{Settles2010Active,Chitta2018Large,Chitta2019Subsampling, Beluch2018Power, Lakshminarayanan2017Simple}, little attention has been given to active learning for DNN-based object detectors, a key ingredient in any autonomous driving system. In addition, none of these active learning for deep object detection works are at the right scale for production-ready systems. 

%%%%%%%%%%%%%%%%%%%%%%%%%%%%%%%%%%%%%%%%%%%%%%%%%%%%%%%%%%%%%%%%%%%%%%%%%%%%%%%%%%%%%%%%%%%%%%%%%%%%%%%%%%%%%%%
%%%%%%%%%%%%%%%%%%%%%%%%%%%%%%%%%%%%%%%%%%%%%%%%%%%%%%%%%%%%%%%%%%%%%%%%%%%%%%%%%%%%%%%%%%%%%%%%%%%%%%%%%%%%%%%

 We have built a scalable production system for active learning in the domain of object detection for autonomous driving. The core of our system is an ensemble of object detectors providing potential bounding boxes and probabilities for each class of interest. Then, a scoring function is used to obtain a single value representing the informativeness of each unlabeled image. Finally, we sample images from the unlabeled pool on the basis of their score and label the selected images. In this paper, we provide a high-level overview of our system as well as a description of the main challenges of building each component.

In contrast to related works in this area, we work with unlabeled data at scale and “in the wild” and we provide an exhaustive set of experiments to compare the different possibilities of each component in our system. Typical object detection research datasets make it possible to simulate the selection from a pool of at most ~80k images (e.g., MS COCO). These images have already been pre-selected for labeling when the dataset was created, and thus contain mostly informative images. In one of our experiments, we select from a pool of 2 million images stemming from recordings collected by cars on the road, so they contain noisy and irrelevant images. A smart selection to drill down to relevant images is absolutely required in this case.

We start by summarizing related works in Section~\ref{sect:sota}. Then, in Section~\ref{sect:method}, we describe the active learning methodology we chose and the different options available for each component. In Section~\ref{sect:experiments}, we present the experiment setup, and discuss our results. Finally, in Section~\ref{sect:abtest}, we describe an active learning A/B test for object detection in a production-ready system and discuss the outcomes.

%%%%%%%%%%%%%%%%%%%%%%%%%%%%%%%%%%%%%%%%%%%%%%%%%%%%%%%%%%%%%%%%%%%%%%%%%%%%%%%%%%%%%%%%%%%%%%%%%%%%%%%%%%%%%%%
%%%%%%%%%%%%%%%%%%%%%%%%%%%%%%%%%%%%%%%%%%%%%%%%%%%%%%%%%%%%%%%%%%%%%%%%%%%%%%%%%%%%%%%%%%%%%%%%%%%%%%%%%%%%%%%
\section{Related Work}
\label{sect:sota}

Active learning aims at finding the minimum number of labeled images to have a supervised learning algorithm reach a certain performance. The main component in an active learning loop is a scoring function which ranks the informativeness of new unlabeled data. That is, if a data point needs to be labeled and added to the training set. Most active learning works are focused on image classification either using classical learning approaches~\cite{Settles2010Active} or deep neural
networks~\cite{Chitta2018Large,Beluch2018Power,Lakshminarayanan2017Simple,Gal2015Dropout,Gal2016Uncertainty,Sener2017Active,atanov2018uncertainty,geifman2018boosting}. For the latter, a powerful and successful approach for deep neural networks to estimate the informativeness of images is based on ensembles~\cite{Beluch2018Power,Lakshminarayanan2017Simple}. Different models in an ensemble of networks are trained independently and are combined to assess the uncertainty of the sample. Ensembles are easy to optimize and fast to execute and have been widely used for classification~\cite{Beluch2018Power,Lakshminarayanan2017Simple,Chitta2018Large}.

There are very few works of deep active learning for object detection. Roy~\etal~\cite{BMVC18_ObjDetec} use a query by committee approach and the disagreement between the convolutional layers in the object detector backbone to query images. Brust~\etal~\cite{VISAPP_ObjDetec} evaluate a set of uncertainty-based active learning aggregation metrics that are suitable for most object detectors. These metrics are focused on the aggregation of the scores associated to the detected bounding boxes. Kao~\etal~\cite{ACCV18_ObjDetec} propose an algorithm that incorporates the uncertainty of both classification and localization outputs to measure how tight the detected bounding boxes are. This approach computes two forward passes, the original image and a noisy version of it, and compares how stable the predictions are. Desai~\etal~\cite{BMVC19_ObjDetec} combine active learning with weakly supervised learning to reduce the efforts needed for labeling. Instead of always querying accurate bounding box annotations, their method first queries a weak annotation consisting of a rough labeling of the object center, and move towards the accurate bounding box annotation only when required. The results on the standard pool-based active learning setting show promising results in terms of the amount of time saved for annotation. More recently, Aghdam~\etal~\cite{ICCV2019_Joost} propose an image-level scoring process to rank unlabeled images for their automatic selection. They first compute the importance of each pixel in the image and aggregate these pixel-level scores to obtain a single image-level score. 

All these methods show promising results using different object detectors on relatively small datasets such as PASCAL VOC and MS-COCO. However, their experiments are focused on the early stage of the active learning process, and therefore consider only small dataset sizes (from 500 up to 3500 images in the case of PASCAL VOC). In general, once the number of training images increases, the improvement of those approaches becomes marginal, as it occurs for instance in~\cite{ACCV18_ObjDetec} with MS-COCO where the training set size reaches 9000 images. In contrast, in this paper, we scale the process starting with at least 100k images and increasing the number of images by 200k in each iteration. Our extensive experiments show the benefits of using active learning for deep object detection within this setting. In addition, we show improvements in a production-like setting, where active learning is used to sample new images from a set of unlabeled data containing more than 2M images.

\section{Scalable Active learning for object detection}
\label{sect:method}

In this section, we describe several active learning techniques that can be applied for object detection and that scale to the dataset sizes we consider in our experiments.

The outline of our active learning framework is shown in~\fig{fig:Active_Learning}. As shown, given a labeled dataset, we initially train one (or more) object detectors which we then use to query from the unlabeled dataset. For querying, we distinguish two phases: scoring and sampling. During scoring, a single individual score per image is computed giving its informativeness. During sampling, a batch of images is selected for labeling, potentially (but not necessarily) making use of the computed informativeness score but also encouraging diversity in the selected batch.

In the next subsections, we first describe a set of scoring functions followed by the sampling strategies we considered.

\subsection{Scoring functions}
\label{sect:scoring}
The goal of the scoring function is to compute a single score per image indicating its informativeness. In this work, we assume the object detector outputs a 2D map of probabilities per class (bicycle, person, car, etc.). Each position in this map corresponds to a patch of pixels in the input image, and the probability specifies whether an object of that class has a bounding box centered there. Such an output map is often found in one-stage object detectors such as SSD \cite{liu2016ssd} or YOLO \cite{redmon2016you}.

In our experiments, we empirically evaluate the following scoring functions:

\textbf{Entropy.} We can compute the entropy of the Bernoulli random variable at each position in the probability map of a specific class. This entropy is computed as follows: 
\begin{equation}
\mathcal{H}(\mathbf{p}_c) = p_c\log p_c + (1 - p_c)\log(1 - p_c),
\end{equation}    
\noindent where $p_c$ represents probability at position $\mathbf{p}$ for class $c$.

\textbf{Mutual Information (MI).}
This method makes use of an ensemble $E$ of models to measure disagreement. First, we compute the average probability over all ensemble members for each position $\mathbf{p}$ and class $c$ as:
\begin{equation}
     \overline{\mathbf{p}}_c =\frac{1}{|E|}\sum_{e\in E} \mathbf{p}_c^{(e)}.
\end{equation}
\noindent where $|E|$ is the cardinality of $E$. Then, the mutual information is computed as:
\begin{equation}
    \mathcal{MI}(\mathbf{p}_c) = \mathcal{H} (\overline{\mathbf{p}}_c) - \frac{1}{|E|}\sum_{e \in E} \mathcal{H} (\mathbf{p}^{(e)}_c), 
\end{equation}
$\mathcal{MI}$ encourages uncertain samples with high disagreement among the ensemble models to be selected during the data acquisition process. 

\textbf{Gradient of the output layer (Grad).} This function measures the uncertainty of the model based on the magnitude of ``hallucinated" gradients \cite{ash2019deep}. Specifically, the predicted label $\tilde{y}$ is assumed to be the ground-truth label $\hat{y}$. Then, the gradient for this label can be computed and its magnitude is used as a proxy for uncertainty. Samples with low uncertainty will produce small gradients. As an extension, by using an ensemble, we associate to each image a single score based on the maximum or mean variance of the gradient vectors.

\textbf{Bounding boxes with confidence (Det-Ent).} We assume the final predicted bounding boxes of our detector have an associated probability. This allows to compute the uncertainty in the form of entropy for each bounding box. We can then apply the same aggregation methods described below to derive a single score per image.

\begin{figure}[t]
 \begin{tabular}{cc}
 \hspace{-0.1cm}\includegraphics[width=0.49\linewidth] {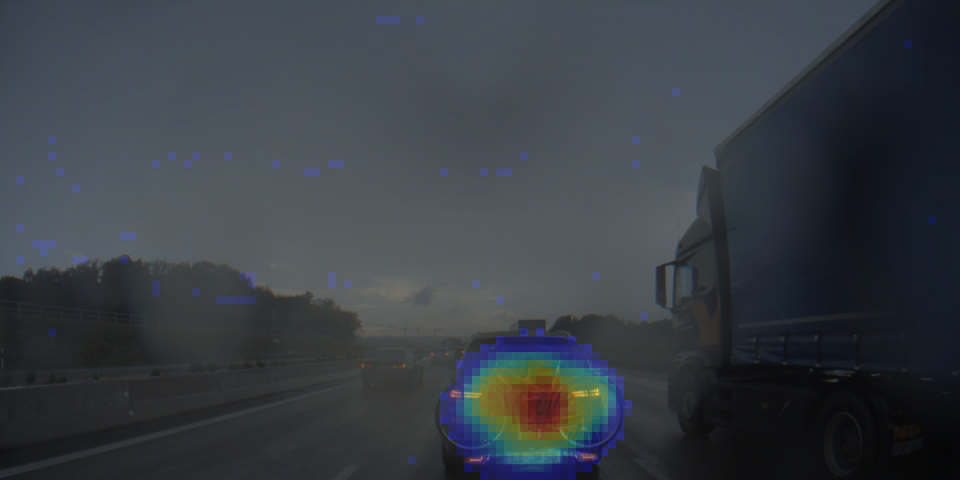}& \hspace{-0.38cm}\includegraphics[width=0.49\linewidth] {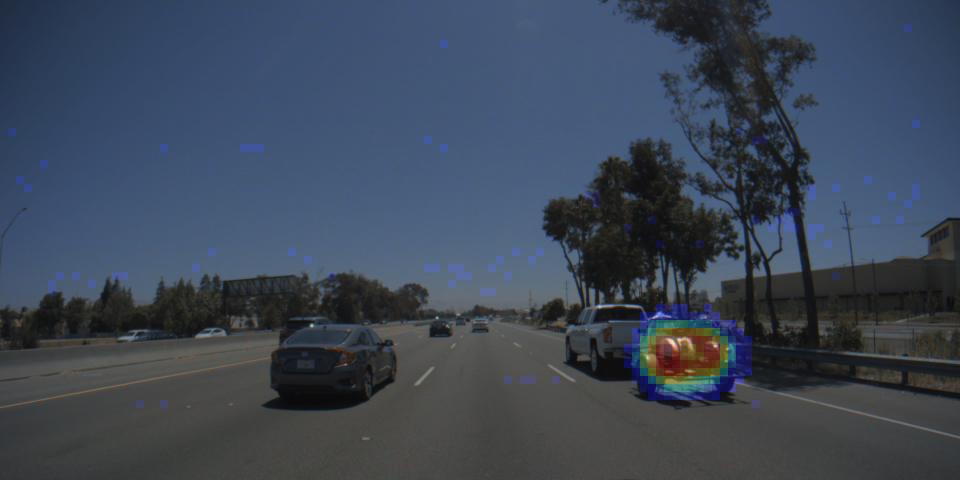}\\ \hspace{-0.2cm}\includegraphics[width=0.49\linewidth] {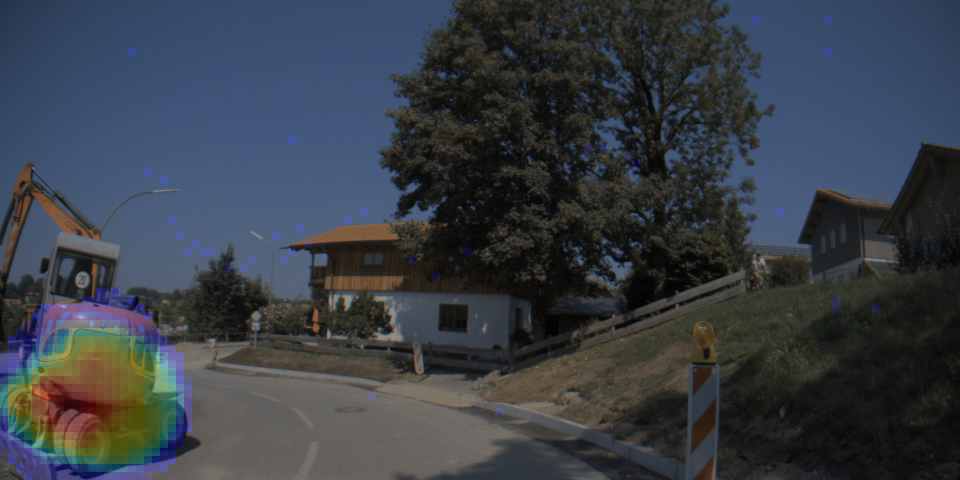}& \hspace{-0.38cm}\includegraphics[width=0.49\linewidth] {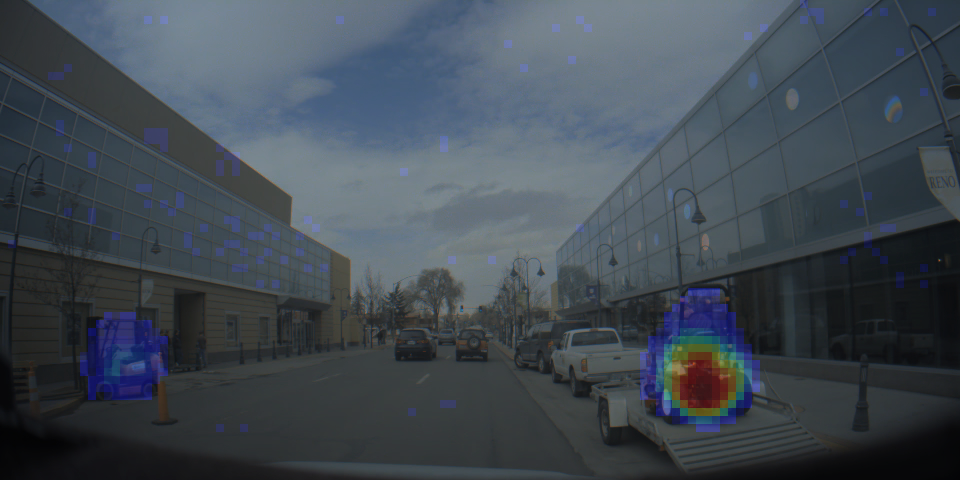}
 \end{tabular}
 %\vspace{-0.25cm}
   \caption{Scoring functions output a 2D-map representing the informativeness score at each pixel and for each class. These scores are then aggregated to provide a single value per image.}
  \label{fig:heatmaps}
  \vspace{-0.5cm}
 \end{figure}

Examples of informativeness maps overlapped on the original image are shown in~\fig{fig:heatmaps}. For a detailed theoretical analysis of these functions, we refer the reader to \cite{Gal2016Uncertainty}.

\subsubsection{\textbf{Score Aggregation}} There are multiple options to aggregate the scores obtained via the techniques above. We experiment with two popular approaches such as taking the maximum or the average\footnote{As an exception, we use sum instead of average, for aggregating entropy when used with \emph{Det-Ent}.}. For the maximum, the score is defined as: %($s=\max_{c \in C} \max_{\mathbf{p}} \mathcal{I}(\mathbf{p}_c)$) 
\begin{equation}
    s=\max_{c \in C} \max_{\mathbf{p}} \mathcal{I}(\mathbf{p}_c).
\end{equation}
Taking the maximum can be prone to outliers (since the probability at a single position will determine the final score), whereas taking the average is biased towards images with many objects (since an image with a single high uncertainty object might get a lower score than an image with many low uncertainty objects).

\subsection{Sampling strategies}
\label{sect:diversity}
In this step, the goal is to select a batch of $N$ samples for annotation from the unlabeled dataset $\mathcal{X}_u$. The most common approach uses only the informativeness score computed in the previous step, for instance, by selecting the top $N$ scoring images for labelling.

One issue with sampling strategies based only on individual image informativeness is that they are prone to rank similar images, e.g. consecutive images in a sequence, similarly high. This potentially wastes training and labeling resources, especially in autonomous driving settings where images come from recorded video sequences. Therefore, the next sampling strategies we consider use the informativeness score from the previous step but in addition (or exclusively) encourage diversity of the selected batch.

For the diversity-based methods, we proceed in three steps: First, extract image embeddings for all the unlabeled samples (i.e., mapping an image to a feature vector). For instance, we use embbedings obtained from the final layer of the backbone. Then, we compute a similarity matrix $D$ based on euclidean distance, $d_{ij}=\norm{e_i - e_j}^2_2$, as well as cosine similarity $d_{ij}={e_i e_j}^{}/{ \norm{e_i}^2_2  \norm{e_j}^2_2 }$.

Based on the similarity matrix, we consider three sampling strategies:  

\textbf{k-means++}~\cite{kmeans++} (KMPP) is an algorihtm used to provide an initialization for the centroids in the popular k-means algorithm~\cite{KMEANS}. This initialization works as follows:
\begin{itemize}
    \item Randomly select the first centroid $c_0$ from the available data $\mathcal{X}$ and add it to $\mathcal{C}$, the centroid set.
    \item For each data point $x\in\mathcal{X}\setminus\mathcal{C}$, compute its distance to the nearest centroid $d_{\text{min}}(x) = \min_{c \in C} d(x, c)$.
    \item Add a new centroid $c_i$ to $\mathcal{C}$ by randomly choosing $x\in\mathcal{X}\setminus\mathcal{C}$ according to a probability distribution proportional to its distance and uncertainty score $s(x)$, \textit{i.e.}, ${p(x) = s(x)d_{\text{min}}(x)} / { \sum_{x\in \mathcal{X} } s(x)d_{\text{min}}(x) }$. That is, the point having maximum distance from the nearest centroid and the highest uncertainty is most likely to be selected next as a centroid.
\end{itemize}
Repeat steps 2 and 3 until $N$ centroids have been sampled.

\textbf{Core-set} (CS) is the subset of data points that best covers the distribution of a larger set of points. In the context of active learning, a core-set approach was initially proposed in~\cite{Sener2017Active}. We leveraged the greedy implementation of their algorithm where the centroid $c_i$ is chosen at each iteration according to:
\begin{equation}
\vspace{-0.1cm}
    c_i=\text{arg}\max_{x\in \mathcal{X}}\min_{c \in \mathcal{C}}s(x)d(x, c).
\end{equation}
Similarly to k-means++, we weight the distance by the uncertainty score of each sample.

\textbf{Sparse Modeling} (OMP)~\cite{OMP} aims at combining uncertainty and diversity using sparse modeling in the sample selection process. More precisely, the algorithm finds a sparse linear combination to represent the uncertainty of unlabeled data where diversity is also incorporated. The goal is to select $N$ samples that can cover the information of the data pool as much as possible. 
\begin{equation}
\begin{array}{rl}
\displaystyle \tilde{x} = \min_{x} & \norm{Dx-s}^2,\\
\textrm{s.t.} & \norm{x}_0 =N,\mathbf{0} \leq x \leq \mathbf{1}. \\
\end{array}
\end{equation}
Intuitively, the optimal solution $\tilde{x}$ is a modified uncertainty score where $N$ scores are re-ranked and the rest are set to zero. In addition, this formulation encourages that samples with high similarity are not selected at the same time: if two samples are similar and both are selected, that would lead to a heavy penalty in the loss function. This optimization problem can be solved in a similar way to orthogonal matching pursuit~\cite{OMPAlg}. Specific details for solving this problem can be found in~\cite{OMP}.

%%%%%%%%%%%%%%%%%%%%%%%%%%%%%%%%%%%%%%%%%%%%%%%%%%%%%%%%%%%%%%%%%%%%%%%%%%%%%%%%%%%%%%%%%%%%%%%%%%%%%%%%%%%%%%%%%%%%%%%%%%%%%%%%%%%%%%%%%
\section{Experiments}
\label{sect:experiments}
In this section, we provide an exhaustive evaluation of different scoring functions and sampling methods. We then compare active learning to random data selection over multiple learning loops. Below, we first detail the experimental settings and then, we discuss our results.

\subsection{Experimental Setup}
For our experiments, we use an internal large scale research dataset consisting of 847k and 33K images for training and testing, respectively. Each image is annotated with up to 5 classes: car, pedestrian, bicycle, traffic sign and traffic light. Unless otherwise specified, we initially train an ensemble of 6 models using a random subset of 100K training images. We use the ensemble to score the remaining training images and select images in batches of $N=$ 200K images. We add the new set of annotated images to the training set so far and train a new model from scratch on this data. We repeat scoring, selecting, and retraining until the entire dataset is selected. Each of the members of the ensemble is a one-stage object detector based on a UNet-backbone. For the evaluation, we consider the performance of a single model and report the weighted mean average precision (wMAP) which averages MAP across several object sizes.%, see~\fig{fig:unet}. For the evaluation, we consider the performance of a single model and report the weighted mean average precision (wMAP) which averages MAP across several object sizes.

%\parbox{\textwidth}{
\begin{table*}[t]
\centering
\makebox[0pt][c]{\resizebox{\textwidth}{!}{%
\begin{minipage}[t]{0.4\textwidth}
\caption{Comparison of scoring functions and score aggregation methods together with the amount of bounding boxes produced by each method. }
\centering
\begin{tabular}{l|c|c|c|} 
\multicolumn{2}{c|}{}&wMAP& \# BBoxes (M)\\ \hline
\multicolumn{2}{l|}{random}&73.36 & 3.70\\ \hline 
\multicolumn{2}{l|}{MC-Dropout~\cite{Gal2015Dropout}}&71.98&3.10 \\ \hline
\multirow{2}{*}{Entropy}&Max.&75.21 & 5.30\\ \cline{2-4}
&Avg.&73.53 & 6.82\\\hline
\multirow{2}{*}{Grad}&Max.&74.98 & 6.42\\ \cline{2-4}
&Avg.&74.72&4.18\\\hline
%\multirow{2}{*}{var}&Max.&74.67\\ \cline{2-3}
%&Avg.&74.30\\ \hline
\multirow{2}{*}{Det-Ent}&Max.&74.40 & 6.90\\ \cline{2-4}
&{Sum.}&\textbf{75.71} & \textbf{7.11}\\\hline
\multirow{2}{*}{MI}&Max.&75.30 & 4.77\\ \cline{2-4}
&Avg.&74.86 &6.03\\
\end{tabular}
\label{tab:scoringfunctions}
 \end{minipage}
     \hfill
    \begin{minipage}[t]{0.21\textwidth}\centering
     \caption{Comparison of sampling strategies using only the informativeness of images. }
    \centering
    \begin{tabular}{l|c|} 
&wMAP\\ \hline
top-$N$&\textbf{75.30}\\
top-third&74.79\\
top-$N/2$-bottom-$N/2$&74.02\\
bottom-$N$&68.01\\
\end{tabular}
\label{tab:sampling2}
     \end{minipage}
    \hfill
    \hspace{0.75cm}\begin{minipage}[t]{0.32\textwidth}\centering
     \caption{Comparison of diversity-promoting sampling strategies. }
    \begin{tabular}{p{0.2cm}l|c|c|c|c|} 
\multicolumn{4}{c|}{}&wMAP\\ \hline
\multicolumn{4}{l|}{random} &69.70\\\hline
\multicolumn{4}{l|}{Top-N} &70.38\\\hline
\multirow{10}{*}{ \begin{sideways}\begin{small}Diversity-promoting\end{small}\end{sideways} }&\multirow{2}{*}{KMPP}&VGG&Eucl.&71.22\\ \cline{3-5}
&&DN&Eucl.&70.17\\ \cline{2-5}
&\multirow{4}{*}{OMP}&\multirow{2}{*}{VGG}&Eucl.&70.82\\ \cline{4-5}
&&&cosine&70.25\\ \cline{3-5}
&&\multirow{2}{*}{DN}&Eucl.&70.71\\\cline{4-5}
&&&cosine&71.32\\ \cline{2-5}
&\multirow{4}{*}{CS}&\multirow{2}{*}{VGG}&Eucl.&71.09\\ \cline{4-5}
&&&cosine&70.90\\\cline{3-5}
&&\multirow{2}{*}{DN}&Eucl.&70.89\\\cline{4-5}
&&&cosine&\textbf{71.33}\\
\end{tabular}
\label{tab:diversitySampling}
    
     \end{minipage}
}}
\vspace{-0.5cm}
\end{table*} 
\subsection{Results}
%%%%%%%%%%%%%%%%%%%%%%%%%%%%%%%%%%%%%%%%%%%%%%%%%%%%%%%%%%%%%%%%%%%%%%%%%%%%%%%%%%%%%%%%%%%%%%%%%%%%%%%%%%%%%%%%%%%%%%%%%%%%%%%%%%%%%%%%%%%%
%%%%%%%%%%%%%%%%%%%%%%%%%%%%%%%%%%%%%%%%%%%%%%%%%%%%%%%%%%%%%%%%%%%%%%%%%%%%%%%%%%%%%%%%%%%%%%%%%%%%%%%%%%%%%%%%%%%%%%%%%%%%%%%%%%%%%%%%%%%%
\subsubsection{Scoring functions} In this experiment, we compare the results of using the five scoring functions defined in Section~\ref{sect:scoring} to random sampling. For each function, we also evaluate the influence of different score aggregation methods. For comparison, we also provide results obtained using MI with MC-Dropout~\cite{Gal2015Dropout}. MC-Dropout obtains multiple predictions based on masks generated in the dropout layers. The main idea is to maintain Dropout active at test time, so each forward pass uses a different dropout-mask, and therefore gives a different prediction. For $T$ forward passes, we can then compute mutual information as coming from an ensemble of $T$ models.

Table~\ref{tab:scoringfunctions} shows the summary of these results. As shown, among functions based only on the confidence map, Entropy using average aggregation and Mutual Information using Max are the ones yielding best performance. Det-Ent, the method combining confidence and bounding boxes outperforms all the others. However, as shown in~\tab{tab:scoringfunctions}, Det-Ent is biased towards images containing large number of objects, and therefore the cost of labeling increases. Taking the labeling cost into account, MI using Max seems to provide the best trade-off.    

%%%%%%%%%%%%%%%%%%%%%%%%%%%%%%%%%%%%%%%%%%%%%%%%%%%%%%%%%%%%%%%%%%%%%%%%%%%%%%%%%%%%%%%%%%%%%%%%%%%%%%%%%%%%%%%%%%%%%%%%%%%%%%%%%%%%%%%%%%%%%%%%%%%%%%%%%
\subsubsection{Data Sampling}
We now focus on analyzing four sampling strategies using only the informativeness score of unlabeled images. In particular, we evaluate \textit{top-N} and \textit{bottom-N} which select the most and least uncertain samples respectively; \textit{top-$N/2$-bottom-$N/2$} which is a combination of both and, \textit{top-third} as a combination between most uncertain and slightly easier samples. %The summary of these results is shown in Table~\ref{tab:sampling2}. 

As shown in Table~\ref{tab:sampling2}, the performance increases as the number of difficult examples in the selection increases. The best performance is obtained selecting only the most uncertain samples (\textit{top-N}). Therefore, for the rest of the experiments, we use \textit{top-N} as reference for a pure score-based sampling strategy.

We now focus on analyzing the sensitivity of one single active learning loop with respect to using diversity-promoting sampling methods. To this end, we consider the same backbone and initial training set as in previous experiments. However, due to the computational limitations of \textit{OMP}, we limit the sampling batch size to $N=10k$. We use two different methods to compute the embeddings of unlabeled data: DN and VGG. The former uses the same object detection backbone used previously. The latter uses a VGG network pre-trained on Imagenet. In both cases, we apply global average pooling on the spatial axes of the final convolutional layer of these backbones to obtain 160D and 512D embeddings for DN and VGG, respectively. We also compare Euclidean and cosine metrics for measuring the distance between two embeddings. 

Table~\ref{tab:diversitySampling} summarizes our results using \textit{KMPP}, \textit{OMP} and \textit{CS} and compares them to \textit{top-N} and \textit{random} sampling. As shown, there are variations in the effect of the metric used (cosine vs Euclidean) although those depend on the method. Nevertheless, in general, adding diversity consistently outperforms random sampling. In addition, diversity used with the appropriated metric outperforms top-N sampling and improves performance by up to 0.95\% and 1.5\% compared to top-N and random sampling, respectively.%However, this improvement is at the expense of an increment in computational cost. \MF{This comment looks rather negative, I would either reformulate in a more neutral way or omit it all along}

\begin{table}[!t]
\caption{Active Learning vs Random over three iterations of the learning loop.}
    \centering
    \resizebox{\columnwidth}{!}{
    \begin{tabular}{l|p{1cm}|c|c|c|c|c|c|c|c|p{1cm}|} 
 \begin{sideways} Iteration\end{sideways}&\begin{sideways} \# Training images\end{sideways}& Method& Data& \begin{sideways} Bicycle\end{sideways}&\begin{sideways} Car\end{sideways} &\begin{sideways} Person\end{sideways} &\begin{sideways} Road Sign\end{sideways} &\begin{sideways} Traffic Light\end{sideways} & wMAP&\begin{sideways}  \# Unique images\end{sideways}\\ \hline
\multirow{3}{*}{1}& \multirow{3}{*}{300k} &\multirow{2}{*}{AL} & $[\mathcal{X}_u \mathcal{X}_l]$ &61.8&91.7&72.0&56.8&71.9&70.8&277k\\\cline{4-11}
& & &$\mathcal{X}_u$ &62.0&91.9&72.5&55.8&72.1&70.9&300k\\ \cline{3-11}
& &random& $\mathcal{X}_u$ & 54.4& 91.9& 68.1& 54.5& 69.5&67.7&300k\\ \hline
\multirow{3}{*}{2}& \multirow{3}{*}{500k}&\multirow{2}{*}{AL} & $[\mathcal{X}_u \mathcal{X}_l]$ &62.6&91.7&73.6&57.9&73.7&71.9&355k\\\cline{4-11}
& & & $\mathcal{X}_u$&61.0&91.8&70.0&57.3&72.4&70.5&500k\\ \cline{3-11}
& &random& $\mathcal{X}_u$& 56.8& 92.2& 68.4& 54.6& 70.5&68.5&500k\\  \hline
\multirow{3}{*}{3}& \multirow{3}{*}{ 700k }&\multirow{2}{*}{AL} & $[\mathcal{X}_u \mathcal{X}_l]$ &\textbf{65.3}&91.9&\textbf{74.7}&\textbf{58.9}&\textbf{75.3}&\textbf{73.2}&402k\\\cline{4-11}
& & & $\mathcal{X}_u$&58.0&91.8&69.1&57.0&71.8&69.5&700k\\ \cline{3-11}
& &random& $\mathcal{X}_u$ & 56.5& 92.3& 69.5& 56.4& 71.2&69.2&700k\\ \hline \hline
 --&~850k & \multicolumn{2}{c|}{full dataset}&55.6&\textbf{92.3}&69.2&56.5&71.6&69.0&850k\\
\end{tabular}
}
\vspace{-0.5cm}
\label{tab:RandomVSAL}
\end{table}

%%%%%%%%%%%%%%%%%%%%%%%%%%%%%%%%%%%%%%%%%%%%%%%%%%%%%%%%%%%%%%%%%%%%%%%%%%%%%%%%%%%%%%%%%%%%%%%%%%%%%%%%%%%%%%%%%%%%%%%%%%%%%%%%%%%%%%%%%%%%%%%%%%%%%

\subsubsection{Active Learning vs Random} In this last experiment, our goal is to analyze the performance of an object detector when it is trained using several consecutive iterations of the learning loop and compare the performance between actively selecting data and random selection. To this end, we use the same backbone as in the previous experiments trained using the same initial 100k labeled images. Then, we iterate 3 times selecting 200k images in each iteration. We use max-MI as acquisition function and, due to computational limitations, we use top-N as sampling strategy.

For active learning (AL), we also consider the case where sampling is performed not only over the unlabeled data $[\mathcal{X}_u]$ but also over the data previously used for training. That is, the acquisition function is applied to $[\mathcal{X}_u \mathcal{X}_l]$. In this case, the selection could lead to repeated samples, but is potentially beneficial for training the model as the number of samples considered difficult for the model is increased, while the annotation costs are reduced as these repeated samples do not need annotation.

\tab{tab:RandomVSAL} shows the summary of our results for this experiment. As a reference, we also include the performance of the model trained using all the data available. As we can see, active learning consistently outperforms random sampling and even improves accuracy compared to training the model with the entire dataset. Interestingly, including previous training data for sampling ($[\mathcal{X}_u \mathcal{X}_l]$) leads to slightly better results with a significant reduction in the number of images to be labeled. For instance, in the third iteration we obtain a 5\% and 5.5\% relative improvement compared to $\mathcal{X}_u$ and random respectively while annotating only 42\% of the data (compared to random).

\section{Active Learning for object detection: an A/B test}
\label{sect:abtest}
We finally focus on evaluating active learning in an autonomous driving production-like setting with a large amount of unlabeled data. Given a reference model, our goal is to find the right training data to improve night-time detection of Vulnerable Road Users (VRU) such as pedestrians and bicycles/motorcycles. Bad illumination and low contrast makes this a challenging setting not only for DNNs but also for the humans in charge of the annotation process. In this experiment, we use our active learning process described above to select the data to be labeled and compare this selection to one done manually by human experts.%\footnote{Manual selection: sample data based on metadata, and then, manually select the most “helpful” frames for the problem at hand. Although this is better than random selection, it is likely to miss data that the model has trouble with. In addition, it is difficult to scale and error-prone}.

\subsection{Experimental setup.} 
We use the same single shot object detection architecture as in the previous experiments, and we train the reference model on the same internal research dataset of 850k images we used in section \ref{sect:experiments}. This dataset contains only few night-time images with bicycle and person exemplars. Then, we use an ensemble of eight models to compute the score on a large pool of unlabeled data. This pool of unlabeled data consists of 2M night-time images shortlisted using acquisition metadata. After applying the acquisition function, we selected the top 19k images in a round-robin fashion over our two classes of interest (person and bicycle).

We compare the active learning selection to a selection done manually of the same size. That is, images are first selected based on metadata (e.g. to find night images), and then, the most “helpful” images are manually selected for the problem at hand by human experts. We expect such a manual selection to be better than a random selection, but it might miss data that the model has trouble with. In addition, it is difficult to scale and error-prone. 

For evaluation, we follow a cross-validation approach to evaluate specifically the performance for bicycle and person at night-time, and repeat the experiments three times. From each newly labeled image set (19k each from manual and active learning selection), we split off (additional) training and test sets in a 90/10 ratio three times.
Below, we report the performance of models trained using the training portion of the split and our initially labeled training data. In addition, we evaluated the newly trained models on the test set used in the previous section, verifying they all provide a similar performance compared to the reference model. This suggests that the initial test set does not capture all possible real-world conditions. 
%%%%%%%%%%%%%%%%%%%%%%%%%%%%%%%%%%%%%%%%%%%%%%%%%%%%%%%%%%%%%%%%%%%%%%%%%%%%%%%%%%%%%%%%%%%%%%%%%%%%%%%%%%%%%%%%%%%%%%%%%%%%%%%%%%%%%%%%%%%%%%%%%%%%%%%%%

\subsection{Results}
Table~\ref{tab:abtest}a shows the relative performance over the reference model on the test data resulting from combining the test sets from manual and actively selected data. As shown, both manual and active learning selection improve over the reference model. That is, adding additional data targeting specific failures leads to performance improvements. For the person class, the data selected by active learning improves weighted mean average precision (wMAP) by 3.3\%, while manually selected data improves only by 1.1\%, which means the relative improvement by AL is 3$\times$. For the bicycle class, the AL-selected data improves wMAP by 5.9\%, compared to an improvement by 1.4\% for manually curated data, i.e., AL-selected data gives a relative improvement of 4.4$\times$. This confirms that AL-selected data performs significantly better on a test set for bicycle and person classes at night, the main goal of this data selection.

\begin{figure}[!t]
 \begin{tabular}{cc}
 \hspace{-0.01cm}\includegraphics[width=0.47\linewidth] {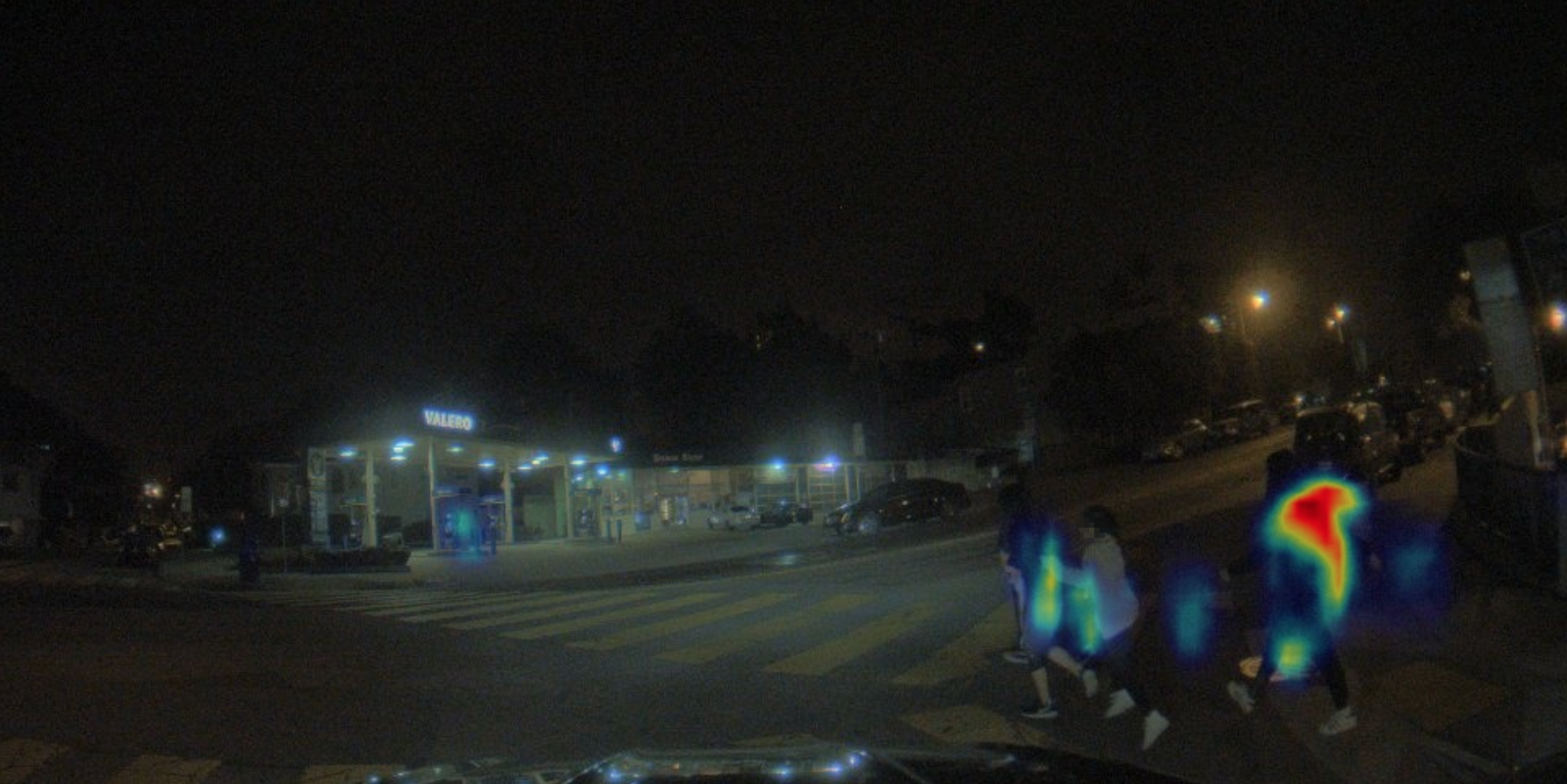}&\hspace{-0.3cm}\includegraphics[width=0.47\linewidth] {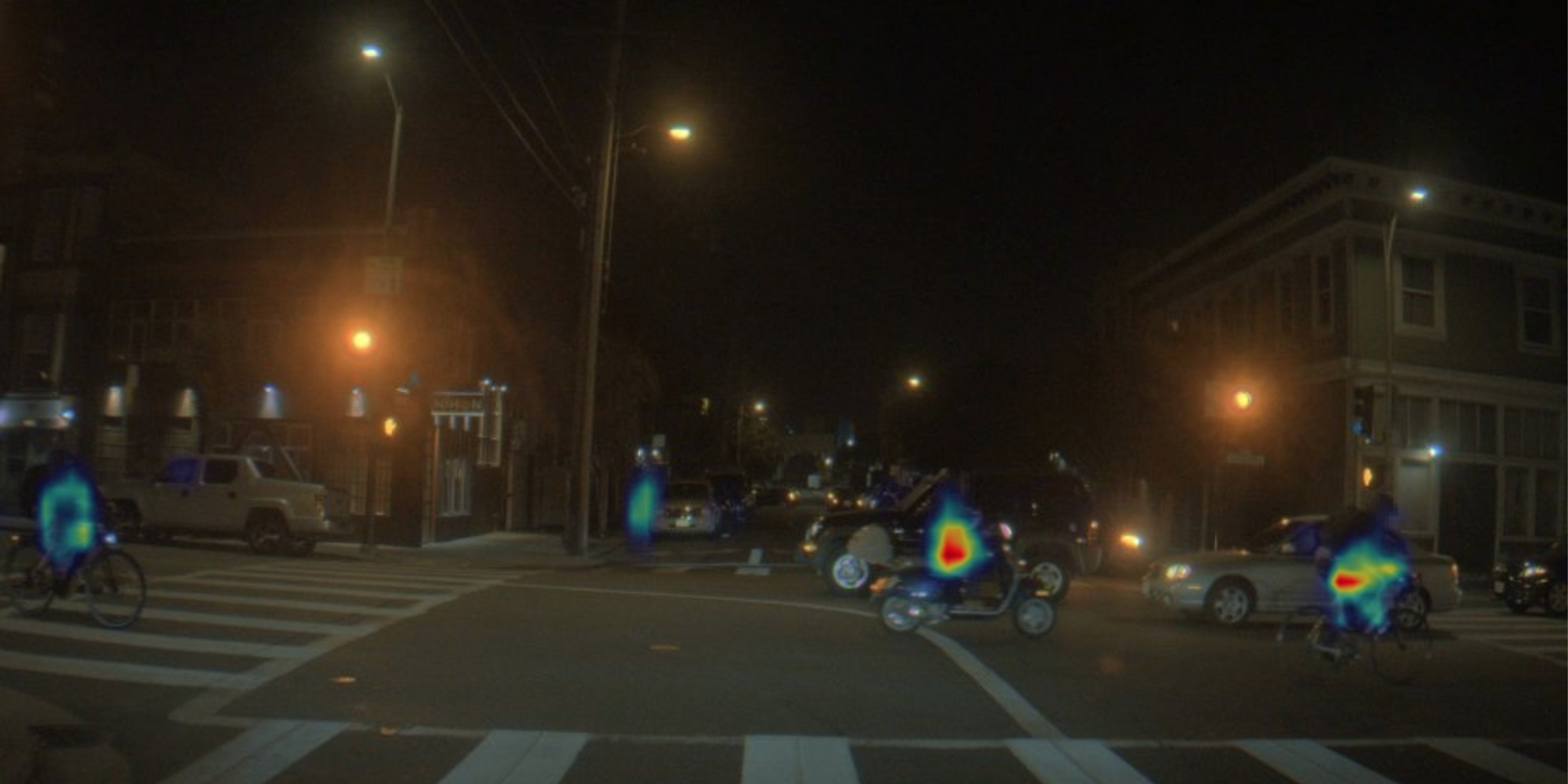}\\
  \hspace{-0.01cm}\includegraphics[width=0.47\linewidth] {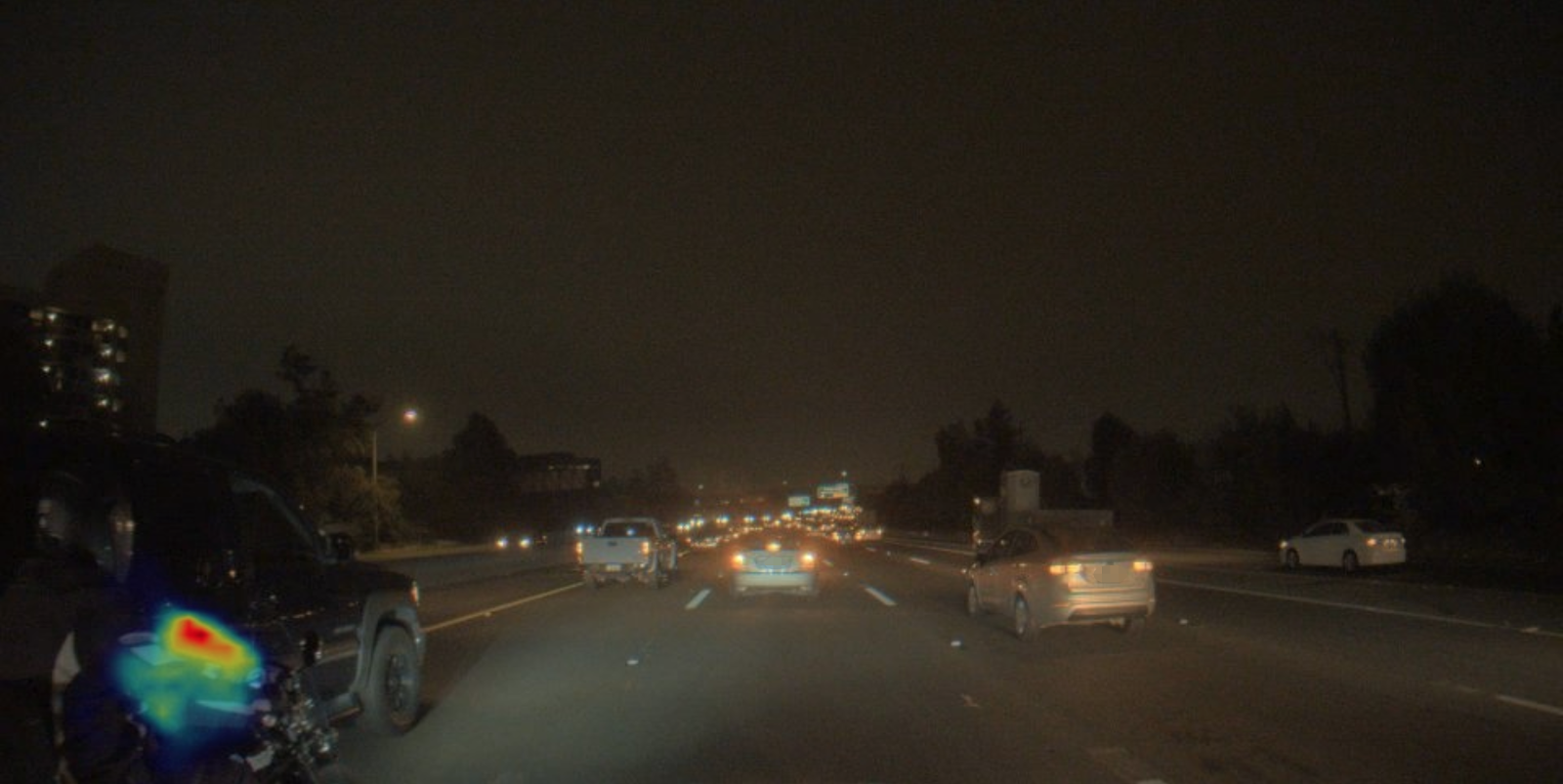}&\hspace{-0.3cm}\includegraphics[width=0.47\linewidth] {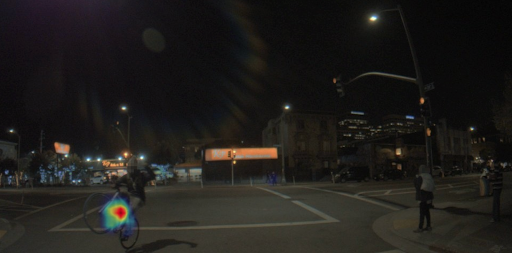}\\
 \end{tabular}
 \vspace{-0.3cm}
  \caption{Representative examples of images selected via active learning. Our system is able to select difficult images for training including low contrast (top-left), hard lighting conditions (top-right), truncation and occlusions (bottom-left), and rare cases (bottom-right).}
  \label{fig:exampleframes}
  \vspace{-0.45cm}
 \end{figure}

In addition, to remove the bias due to the selection method, we provide the evaluation of both methods using only test data from the manual selection process. Results of this comparison are summarized in~Table~\ref{tab:abtest}b. In this case, for bicycle objects, the improvement due to active learning data is significantly higher (3.2\% vs 2.3\%), while performing on par for persons and cars. These results demonstrate that the data selected by active learning is at least as good as manually curated data, even when testing exclusively on manually selected data. Qualitative examples of selected images using our active learning approach are shown in~\fig{fig:exampleframes}. As shown, these selected frames are highly informative, while manual selection typically resort to selecting sub-sequences of many images within a video sequence.
\begin{table}[!t]
    \centering
    \caption{Mean average precision evaluated on: a) both manual and active learning test data. b) manual test data only.}
    \vspace{-0.15cm}
    \begin{tabular}{l|c|c||c|c|}
&\multicolumn{2}{c|}{Manual and AL test data}& \multicolumn{2}{|c|}{Manual test data only}\\ \cline{2-5}
&Manual   & Active Learning & Manual   & Active Learning \\ \hline 
     bicycle & +1.36 & \textbf{+5.94} & +2.28 &  \textbf{+3.18}  \\ \hline 
     person &+0.74 &\textbf{+3.28}& +1.28 &   \textbf{+1.35} \\ \hline 
     car & +0.62 &\textbf{+1.07}& \textbf{+0.58} &  +0.28  \\ \hline 
     \multicolumn{1}{c}{}&\multicolumn{2}{c}{ (a) }& \multicolumn{2}{c}{ (b) }
    \end{tabular}
      \label{tab:abtest}
\vspace{-0.55cm}
\end{table}

Both methods selected the same amount of data for labeling. Interestingly, the labeling costs for both methods are similar (within 5\%, measured via annotation time). However, comparing the number of objects in each dataset, we observe that the number of objects selected by our active learning approach is around 12\% higher. The selection done via active learning also contains more person and bicycle objects, while fewer car and other objects. Therefore, we can conclude that the data selected by our approach is more directed to the problem at hand. 

From these results, we can conclude that our approach shows a strong improvement from data selected via active learning compared to a manual curation by experts.

\section{Conclusions}
Collecting and annotating the right data for supervised learning is an expensive and challenging task that can significantly improve the performance of any perception-based autonomous driving system. In this paper, we described our scalable production system for active learning for object detection. The main component of this system is an image-level scoring function that evaluates the informativeness of each new unlabeled image. Then, selected images are labeled and added to the training set. We first provided a comprehensive comparison of different scoring and sampling methods, and then used our system to improve the accuracy for night-time and vulnerable road users in a production-like setting. Our experimental results show very strong performance improvements for the automatic selection, with up to 4$\times$ the relative mean average precision improvement compared to a manual selection process by experts.

{\small
\bibliographystyle{IEEEtranBST/IEEEtran}
\bibliography{bib}

\begin{thebibliography}{10}
\providecommand{\url}[1]{#1}
\csname url@rmstyle\endcsname
\providecommand{\newblock}{\relax}
\providecommand{\bibinfo}[2]{#2}
\providecommand\BIBentrySTDinterwordspacing{\spaceskip=0pt\relax}
\providecommand\BIBentryALTinterwordstretchfactor{4}
\providecommand\BIBentryALTinterwordspacing{\spaceskip=\fontdimen2\font plus
\BIBentryALTinterwordstretchfactor\fontdimen3\font minus
  \fontdimen4\font\relax}
\providecommand\BIBforeignlanguage[2]{{%
\expandafter\ifx\csname l@#1\endcsname\relax
\typeout{** WARNING: IEEEtran.bst: No hyphenation pattern has been}%
\typeout{** loaded for the language `#1'. Using the pattern for}%
\typeout{** the default language instead.}%
\else
\language=\csname l@#1\endcsname
\fi
#2}}

\bibitem{Settles2010Active}
B.~Settles, ``Active learning literature survey,'' Tech. Rep., 2010.

\bibitem{Chitta2018Large}
K.~{Chitta}, J.~M. {Alvarez}, and A.~{Lesnikowski}, ``{Large-Scale Visual
  Active Learning with Deep Probabilistic Ensembles},''
  \emph{arXiv:1811.03575}, 2018.

\bibitem{Chitta2019Subsampling}
K.~Chitta, J.~M. Alvarez, E.~Haussmann, and C.~Farabet, ``{Less is more: An
  exploration of data redundancy with active dataset subsampling},''
  \emph{arXiv:1811.03542}, 2019.

\bibitem{Beluch2018Power}
W.~H. Beluch, T.~Genewein, A.~Nurnberger, and J.~M. Kohler, ``The power of
  ensembles for active learning in image classification,'' in \emph{CVPR},
  2018.

\bibitem{Lakshminarayanan2017Simple}
B.~Lakshminarayanan, A.~Pritzel, and C.~Blundell, ``Simple and scalable
  predictive uncertainty estimation using deep ensembles,'' in \emph{NIPS},
  2017.

\bibitem{Gal2015Dropout}
Y.~{Gal} and Z.~{Ghahramani}, ``{Dropout as a Bayesian Approximation:
  Representing Model Uncertainty in Deep Learning},'' \emph{arXiv e-prints}, p.
  arXiv:1506.02142, Jun 2015.

\bibitem{Gal2016Uncertainty}
Y.~Gal, ``Uncertainty in deep learning,'' Ph.D. dissertation, University of
  Cambridge, 2016.

\bibitem{Sener2017Active}
O.~{Sener} and S.~{Savarese}, ``{Active Learning for Convolutional Neural
  Networks: A Core-Set Approach},'' in \emph{ICLR}, 2018.

\bibitem{atanov2018uncertainty}
A.~{Atanov}, A.~{Ashukha}, D.~{Molchanov}, K.~{Neklyudov}, and D.~{Vetrov},
  ``{Uncertainty Estimation via Stochastic Batch Normalization},'' \emph{ArXiv
  e-prints}, 2018.

\bibitem{geifman2018boosting}
Y.~{Geifman}, G.~{Uziel}, and R.~{El-Yaniv}, ``{Bias-Reduced Uncertainty
  Estimation for Deep Neural Classifiers},'' \emph{ICLR}, 2019.

\bibitem{BMVC18_ObjDetec}
S.~Roy, A.~Unmesh, and V.~P. Namboodiri, ``Deep active learning for object
  detection,'' in \emph{BMVC}, 2018.

\bibitem{VISAPP_ObjDetec}
C.-A. Brust, C.~Käding, and J.~Denzler, ``Active learning for deep object
  detection,'' in \emph{VISAPP}, 2019.

\bibitem{ACCV18_ObjDetec}
C.-C. Kao, T.-Y. Lee, P.~Sen, and M.-Y. Liu, ``Localization-aware active
  learning for object detection,'' in \emph{ACCV}, 2018.

\bibitem{BMVC19_ObjDetec}
S.~V. Desai, A.~C. Lagandula, W.~Guo, S.~Ninomiya, and V.~N. Balasubramanian,
  ``An adaptive supervision framework for active learning in object
  detection,'' in \emph{BMVC}, 2019.

\bibitem{ICCV2019_Joost}
H.~H. Aghdam, A.~Gonzalez-Garcia, J.~van~de Weijer, and A.~M. Lopez, ``Active
  learning for deep detection neural networks,'' in \emph{ICCV}, 2019.

\bibitem{liu2016ssd}
W.~Liu, D.~Anguelov, D.~Erhan, C.~Szegedy, S.~Reed, C.-Y. Fu, and A.~C. Berg,
  ``Ssd: Single shot multibox detector,'' in \emph{ECCV}, 2016.

\bibitem{redmon2016you}
J.~Redmon, S.~Divvala, R.~Girshick, and A.~Farhadi, ``You only look once:
  Unified, real-time object detection,'' in \emph{CVPR}, 2016.

\bibitem{ash2019deep}
J.~T. Ash, C.~Zhang, A.~Krishnamurthy, J.~Langford, and A.~Agarwal, ``Deep
  batch active learning by diverse, uncertain gradient lower bounds,''
  \emph{arXiv:1906.03671}, 2019.

\bibitem{kmeans++}
D.~Arthur and S.~Vassilvitskii, ``K-means++: The advantages of careful
  seeding,'' in \emph{ACM-SIAM SODA}, 2007.

\bibitem{KMEANS}
J.~B. MacQueen, ``Some methods for classification and analysis of multivariate
  observations,'' in \emph{Proceedings of 5-th Berkeley Symposium on
  Mathematical Statistics and Probability}, 1967.

\bibitem{OMP}
G.~{Wang}, J.~{Hwang}, C.~{Rose}, and F.~{Wallace}, ``Uncertainty-based active
  learning via sparse modeling for image classification,'' \emph{IEEE Trans. on
  Image Processing}, vol.~28, no.~1, pp. 316--329, Jan 2019.

\bibitem{OMPAlg}
Y.~C. Pati, R.~Rezaiifar, and P.~S. Krishnaprasad, ``Orthogonal matching
  pursuit: Recursive function approximation with applications to wavelet
  decomposition,'' in \emph{27th Asilomar Conf. Signals, Syst. Comput.}, 1993.

\end{thebibliography}
}

\end{document}